\def\year{2020}\relax
\documentclass[letterpaper]{article} 
\usepackage{aaai20}  
\usepackage{times}  
\usepackage{xcolor}
\usepackage{helvet} 
\usepackage{courier}  
\usepackage[hyphens]{url}  
\usepackage{graphicx} 
\usepackage{graphicx}  
\title{``To Target or Not to Target": Identification and Analysis of Abusive Text Using
Ensemble of Classifiers }

\author{Gaurav Verma, Niyati Chhaya, Vishwa Vinay\\
\{gaverma, nchhaya, vinay\}@adobe.com\\
Adobe Research, India
}

\begin{document}

\maketitle

\begin{abstract}
\vspace{-1mm}
With rising concern around abusive and hateful behavior on social media platforms, we present an ensemble learning method to identify and analyze the linguistic properties of such content. Our stacked ensemble comprises of $3$ machine learning models that capture different aspects of language and provide diverse and coherent insights about inappropriate language. The proposed approach provides comparable results to the existing state-of-the-art on the Twitter Abusive Behavior dataset \cite{founta2018large} \textit{without} using any user or network-related information; solely relying on textual properties. We believe that the presented insights and discussion of shortcomings of current approaches will highlight potential directions for future research.
\end{abstract}
\vspace{-5mm}
\section{Introduction}
Inappropriate language on social media has raised concerns among the users as well as the moderators. Multiple manifestations of online inappropriateness, involving abuse, hate speech, sexism, racism, cyberbullying, and harassment, further amplifies the problem of its categorization given the complex interactions between each of these categories. This has encouraged a lot of research on the topic, including defining the multiple ``faces'' of inappropriateness and curating crowdsourced datasets \cite{founta2018large}, building machine learning (ML) models to identify abusive and hateful content \cite{founta2019unified}, and highlighting the risk of deploying such models trained on biased datasets \cite{sap2019risk}. An important direction that is relatively less explored is the analysis of inappropriate content that is facilitated by the learned patterns of various ML models. We believe that this is an important direction because the ML models can not only identify inappropriate language, but can also aid in highlighting aspects of language that make it so. The understanding of these aspects can be used to further improve systems to perform more robust and bias-free identification.

In this work we primarily focus on two aspects relating to inappropriate language: (1) training ML models to \textit{identify} online inappropriateness, (2) \textit{analyzing} the learnings of trained ML models to gain insights into inappropriate language. Furthermore, we believe that some of our findings can give insights into possible directions for future research. 

In spirit of the Task 2 of ICWSM'20 Data Challenge, we conduct experiments on the Twitter Abusive Behavior dataset \cite{founta2018large} wherein we consider four categories -- normal, spam, abusive, and hateful. We train an ensemble of three sufficiently diverse machine learning models to classify tweets into these categories: (a) a logistic regression classifier on pyscholinguistic features, (b) a bag of n-gram based classifier, and (c) an attention-based bidirectional LSTM classifier. The choice of these models is governed by their ability to provide not only good identification/classification results, but also interpretable insights based on their learned parameters. We then concatenate the predictions of these three models to learn a stacked ensemble (again, a logisitic regression classifier). Our models give results on par with the state-of-the-art \cite{founta2019unified} \textit{without} using network or user metadata; solely relying on linguistic properties contained within the tweet. In the following sections we discuss these classification models and the associated insights. We end with a discussion about the difficulty of learning subtle differences in abusive and hateful tweets, while giving some possible future directions. 

\vspace{-2mm}
\section{Classification Experiments}

We conduct classification experiments on Twitter Abusive Behavior dataset \cite{founta2018large}. The dataset comprises of $\sim100,000$ tweets classified into four categories: normal $(53.85 \%)$, spam $(27.15 \%)$, abusive $(14.04 \%)$, and hateful $(4.96 \%)$. We preprocess the tweets by converting all letters to lowercase. To limit the vocabulary size, we drop the hashtag symbol (\#) while keeping the following tag-word as it often carries crucial information. We replace all user mentions with a common token (e.g., \texttt{@TheTweetOfGod} is replaced by \texttt{user\_tag}); similarly web links are replaced by the token \texttt{web\_link}. Additionally, we remove all  non-alphanumeric characters from the tweets (except spaces and punctuation). Following this preprocessing, we train machine learning models to solve the multiclass classification task. For training and validating the models, we split the entire dataset into train, validation, and test sets in $0.8:0.1:0.1$ ratio. For consistency, these sets are kept the same while training and evaluating all the models. Next, we discuss the models and the insights they offer. 

\subsection{Logistic Regression on Linguistic Features}
\begin{figure*}[!h]
    \centering
    \begin{minipage}{0.23\textwidth}
        \centering
        \fbox{\includegraphics[width=1.0\textwidth]{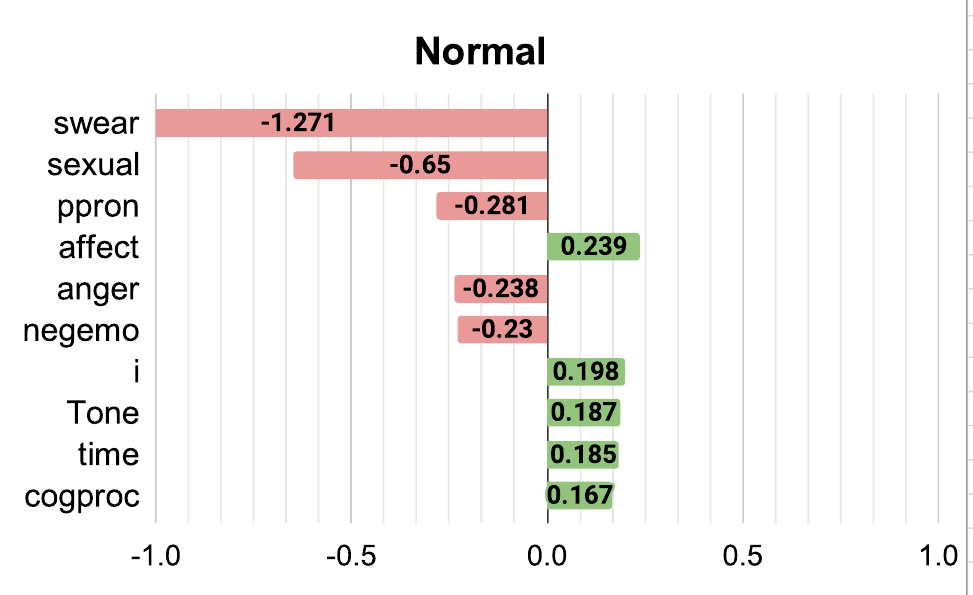}}
    \end{minipage}\hspace{2mm}
    \begin{minipage}{0.23\textwidth}
        \centering
        \fbox{\includegraphics[width=1.0\textwidth]{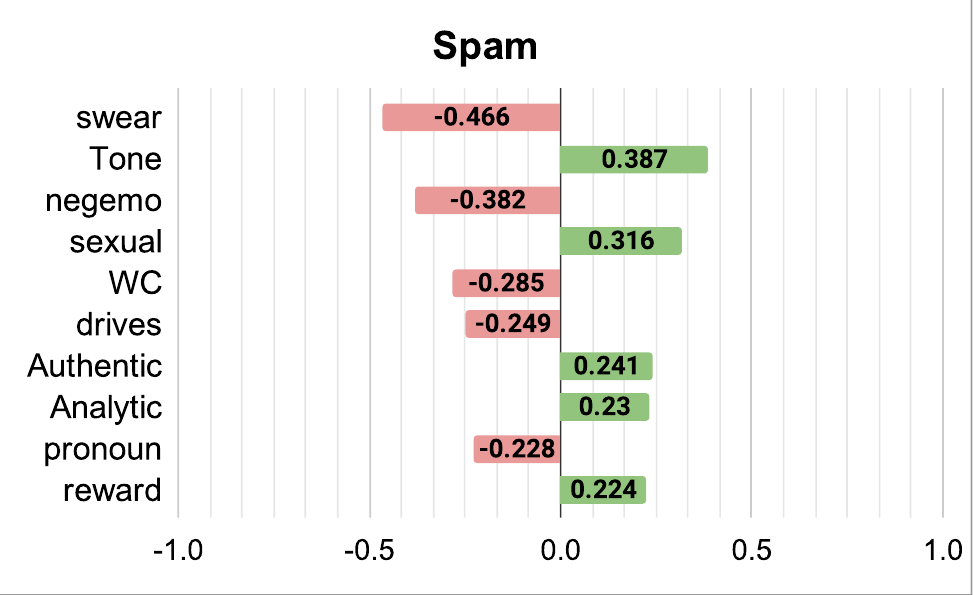}}
    \end{minipage}\hspace{2mm}
    \begin{minipage}{0.23\textwidth}
        \centering
        \fbox{\includegraphics[width=1.0\textwidth]{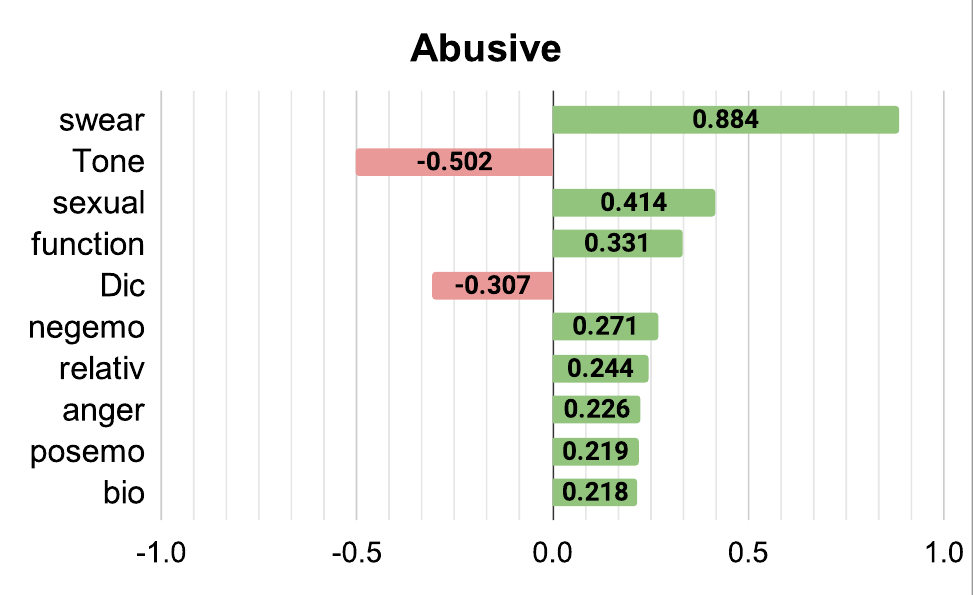}}
    \end{minipage}\hspace{2mm}
    \begin{minipage}{0.23\textwidth}
        \centering
        \fbox{\includegraphics[width=1.0\textwidth]{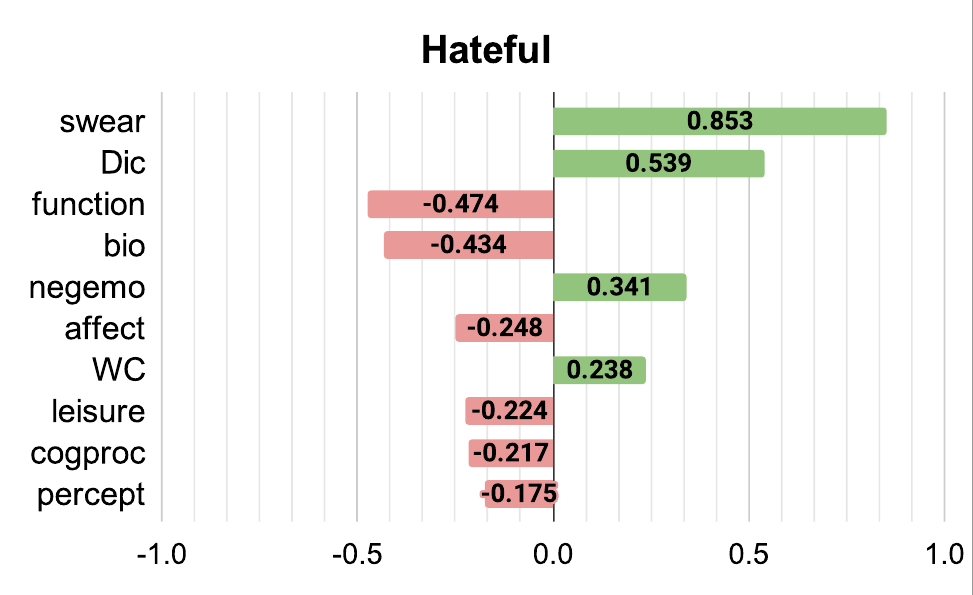}}
    \end{minipage}
    \caption{Logistic Regression on LIWC features: for each class, we depict the top-10 learned coefficients based on their absolute values and the corresponding features. The figure is best viewed on screen with zoom.}
    \label{fig:coeff_analysis}
\end{figure*}

LIWC \cite{liwcPennebaker} is a text analysis software\footnote{https://liwc.wpengine.com/} that allows categorization of words into psychologically meaningful categories. Prior works have demonstrated its ability to capture aspects related to ``attentional focus, emotionality, social relationships, thinking styles, and individual differences" expressed in language \cite{tausczik2010psychological}. We use all the $94$ scores obtained for each tweet as its feature representation to train a logistic regression (LR) model. We standardize the input data and remove highly correlated features (i.e., pearson correlation coefficient $> 0.9$). In Figure \ref{fig:coeff_analysis} we show the top-10 learned coefficients based on their absolute values.\\

From Figure \ref{fig:coeff_analysis} we can infer that personal pronouns (e.g., I, me, mine, you, we) are not strong indicators of spam, abusive, or hateful content whereas they occur frequently in normal content. Similar observation holds true for words indicating time or duration (until, now, season). This shows that text related to abusive behavior tends to be aloof from the author, the ownership or specifics are avoided in order to disassociate themselves from the message. Abusive and hateful content have a high concentration of swear words, as well as express different forms of affect (use of words such as `happy', `cried', `hurt',`ugly',`nasty') and emotions (tone), in turn indicating expressions and opinions. Hateful language uses well-formed words (dic) whereas abusive tweets allude to the use of ill-formed jargon. This composition indicates that expressive, well-formed content may typically be hateful, whereas expressive emotional~(tone) content is likely to be abusive. From Table \ref{tab:classification_results} we see that LR gives a decent classification accuracy; Figure \ref{fig:confusion_matrix} presents the confusion matrix for a closer look at label-wise predictions. 

\bgroup
\def\arraystretch{1.1}
\begin{table}[!t]
    \centering
    \begin{tabular}{ l | c  }
        \textbf{Model} & \textbf{Accuracy} \\\hline
        LR on LIWC features & $0.78$\\
        N-gram based Classification & $0.80 $  \\
        Attention-based BiLSTM & $0.81 $ \\
        Stacked Ensemble & $0.83 $ \\
    \end{tabular}
    \vspace{-2mm}
    \caption{Classification accuracy on the test set. }
    \label{tab:classification_results}
    \vspace{-4mm}
\end{table}
\egroup

\vspace{-2mm}
\subsection{N-gram based Classification}
\citeauthor{joulin2017bag} (\citeyear{joulin2017bag}) proposed fastText -- a simple and efficient baseline for text classification that uses bag of n-gram features to capture partial information about the local word order. We use the fastText library\footnote{https://github.com/facebookresearch/fastText} to train a classifier for our task. We train the model for $10$ epochs with a learning rate initialized at $0.1$. We set the maximum length of n-grams to $3$. The trained model provides vector representations (or embeddings) of the sentences as well as for the words in the vocabulary. The word embeddings allow us to execute nearest neighbor queries as well as perform analogy operations. For instance, we find that the nearest neighbors for offensive words like `fu*king' or `fu*k' are also offensive, yet \textit{diverse}\footnote{For reference, word2vec \cite{mikolov2013distributed}, widely used word embeddings, lists fu@kin, f\_ck, f\_*\_cking, friggin, freakin, fu@ked, (censoring by \textit{us}, in \textit{this} \textit{particular} list, is done by using `@' symbol) etc. as nearest neighbors of the word fu*king. }, in nature -- a*sholes, bullsh*t, su*ks, pen*s, dumba*s, sh*tty, etc.
We believe that the nearest neighbor querying that this approach enables can be used to expand on the dictionary of offensive words. Furthermore, we observe interesting word-level analogies like (word2vec \cite{mikolov2013distributed} responses are given in parentheses for reference):\\
\textit{(a)} fu*king $-$ abuse $+$ normal $=$ boring (w2v: f\_**\_king)\\
\textit{(b)} fata*s $-$ hate $+$ normal $=$ pathetic (w2v: sh*thead)\\
\textit{(c)} b*tch $-$ hate $+$ normal $=$ nasty (w2v: haters)\\

\begin{figure*}[!h]
    \centering
    \begin{minipage}{0.23\textwidth}
        \centering
        \fbox{\includegraphics[width=1.0\textwidth]{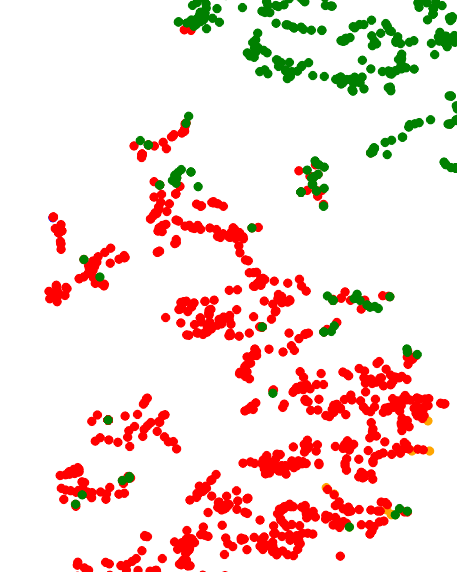}}
    \end{minipage}\hspace{2mm}
    \begin{minipage}{0.57\textwidth}
        \centering
        \includegraphics[width=1.0\textwidth]{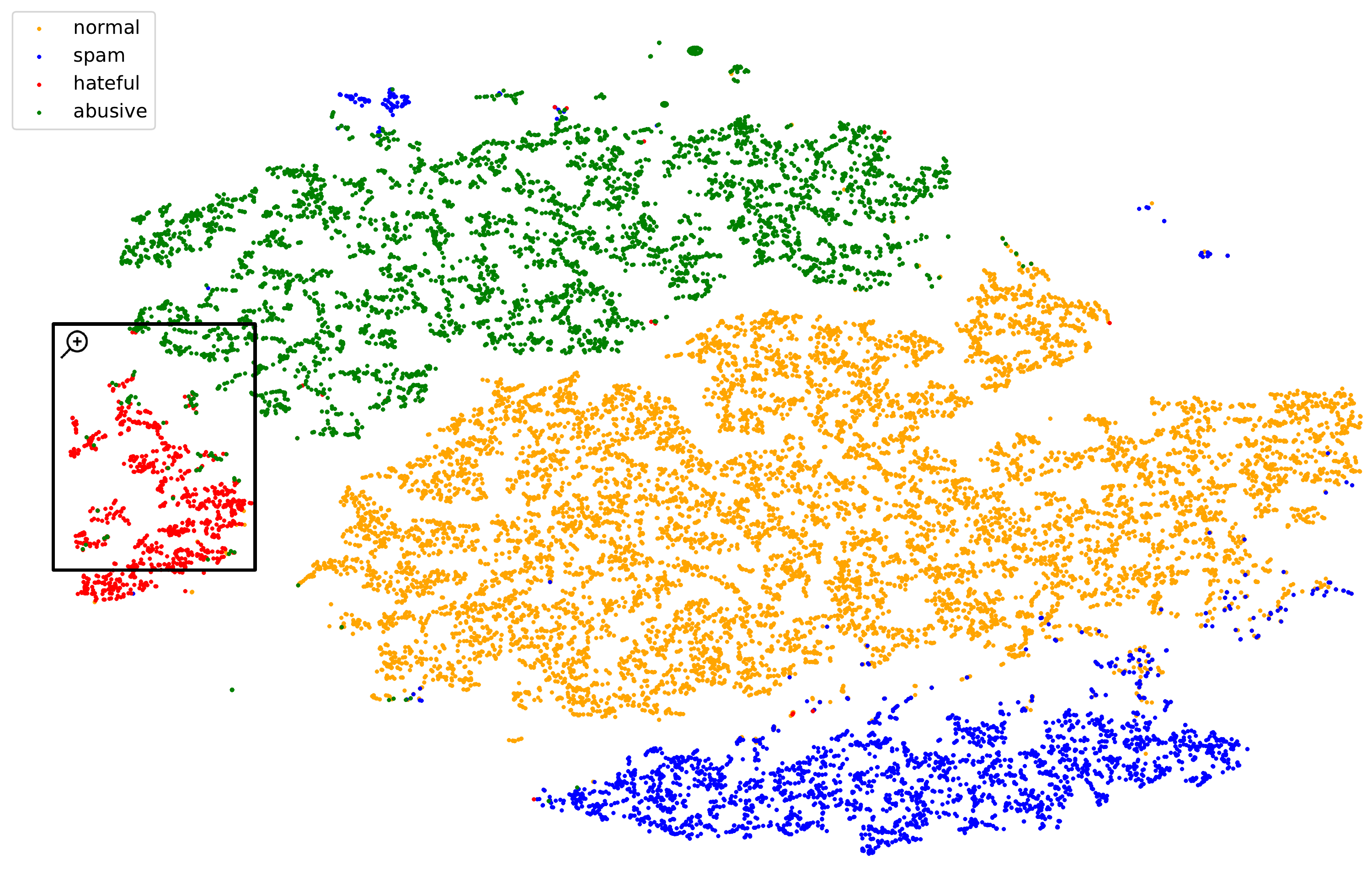}
    \end{minipage}
    \vspace{-2mm}
    \caption{t-SNE plot of sentence embeddings obtained from the n-gram based classifier. In inset, we show that many abusive tweets have similar embeddings as hateful tweets. }
    \label{fig:tsne_plot}
    \vspace{-4mm}
\end{figure*}

 In absence of the context these words are used in, it is difficult to interpret these analogies. However, there is a clear shift \textit{away} from inappropriate expression \textit{toward} more acceptable words while preserving the broad meaning.
 These analogous words can have a potential use in suggesting ``milder'' words to the users as they express themselves on digital platforms or to do counterfactual modeling of abusive and hateful tweets. 
 In Figure \ref{fig:tsne_plot} we show the t-SNE plot of tweet embeddings obtained from the trained model. In inset, we highlight the tendency of this model to classify hateful tweets as abusive tweets; an observation that is reasserted by the numbers presented in the confusion matrix in Figure \ref{fig:confusion_matrix}. Given that this tendency is consistent across all the models under consideration, we revisit this observation later. 

\bgroup
\def\arraystretch{1.05}
\begin{table}[!t]
    \centering
    \begin{tabular}{ l | l | l | l }
        \textbf{Normal} & \textbf{Spam} & \textbf{Abusive} & \textbf{Hateful} \\\hline
        business & hoodies & {jack*ss} & ret*rds \\
        gather & {advertise} & {fu*king} & spitt*ng \\
        snapped & online & bruh & {n*zi} \\
        holds & store & di*khead & ch*ke \\
        {freaking} & horoscopes & fat*ss & b*tch \\
    \end{tabular}
    \vspace{-2mm}
    \caption{Some of the most attended words for each class by the attention-based BiLSTM.}
    \label{tab:attention_weights}
    \vspace{-4mm}
\end{table}
\egroup

\vspace{-2mm}
\subsection{Attention-based Bidirectional LSTM}
\citeauthor{zhou-etal-2016-attention-based} (\citeyear{zhou-etal-2016-attention-based}) proposed a bidirectional LSTM model with attention. Being bidirectional in nature, the model can encode both the left and right sequence context in natural language. The attention module allows the model to ``attend" to input words while performing classification or generation tasks. The attention weights are often used to interpret which input words were crucial for a given prediction -- higher the attention weight, larger is the contribution of that word toward the prediction. Owing to the remarkable modeling capabilities of LSTMs and interpretability of attention module, we train the attention-based BiLSTM model to perform our classification task. We initialize the word embeddings using GloVe representations trained on Twitter corpus \cite{pennington2014glove} and train the model\footnote{https://github.com/TobiasLee/Text-Classification} for $4$ epochs with a learning rate initialized at $10^{-3}$. Following training, we compute the class-wise average of attention weights for all the words in train, validation, and test examples.
Table \ref{tab:attention_weights} shows some of the ``highly attended" words for each of the $4$ classes and gives an estimate of the words that are considered important by the model for making a certain prediction -- in our case, classifying into one of the $4$ categories. For instance, `freaking' -- a euphemism for `fu*king' is considered important for classifying tweets as \texttt{normal}. Even though the words under \texttt{abusive} and \texttt{hateful} categories seem similar in nature, there are certain subtle differences. \citeauthor{founta2018large} (\citeyear{founta2018large}) claim that hateful tweets often contain a well-defined description of the target group(s) whereas abusive tweets do not. This is reflected in Table \ref{tab:attention_weights} as words like `ret*rds' and `n*zi' are often used to target groups unlike words like `jacka*s' and `fata*s'. Interestingly, the words under \texttt{spam} category are also often encountered in advertisements and posts by Twitter bots. 

\vspace{-2mm}
\subsection{Stacked Ensemble}

\begin{figure*}[!h]
    \centering
    \begin{minipage}{0.23\textwidth}
        \centering
        {\includegraphics[width=1.0\textwidth]{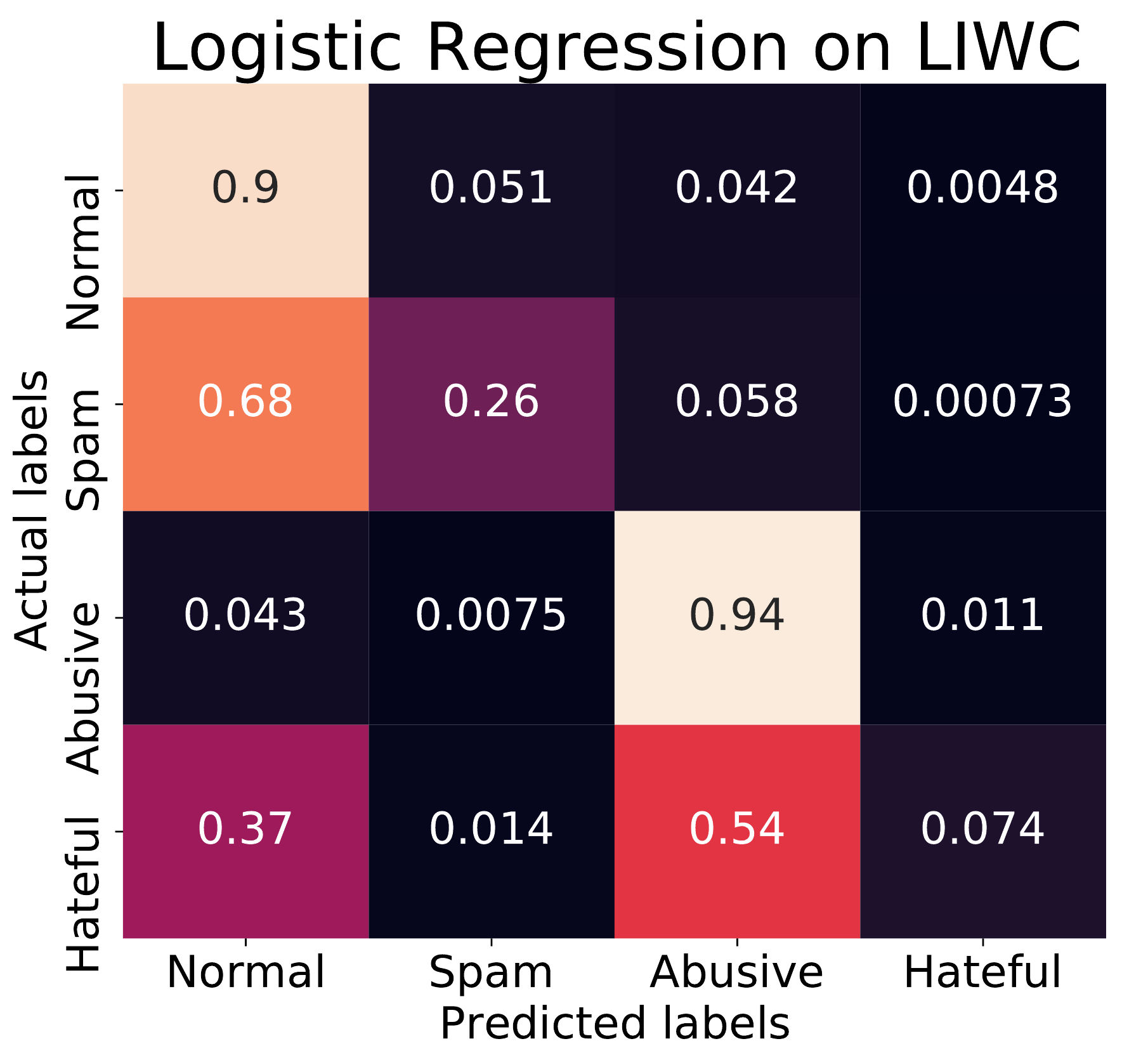}}
    \end{minipage}\hspace{2mm}
    \begin{minipage}{0.23\textwidth}
        \centering
        \includegraphics[width=1.0\textwidth]{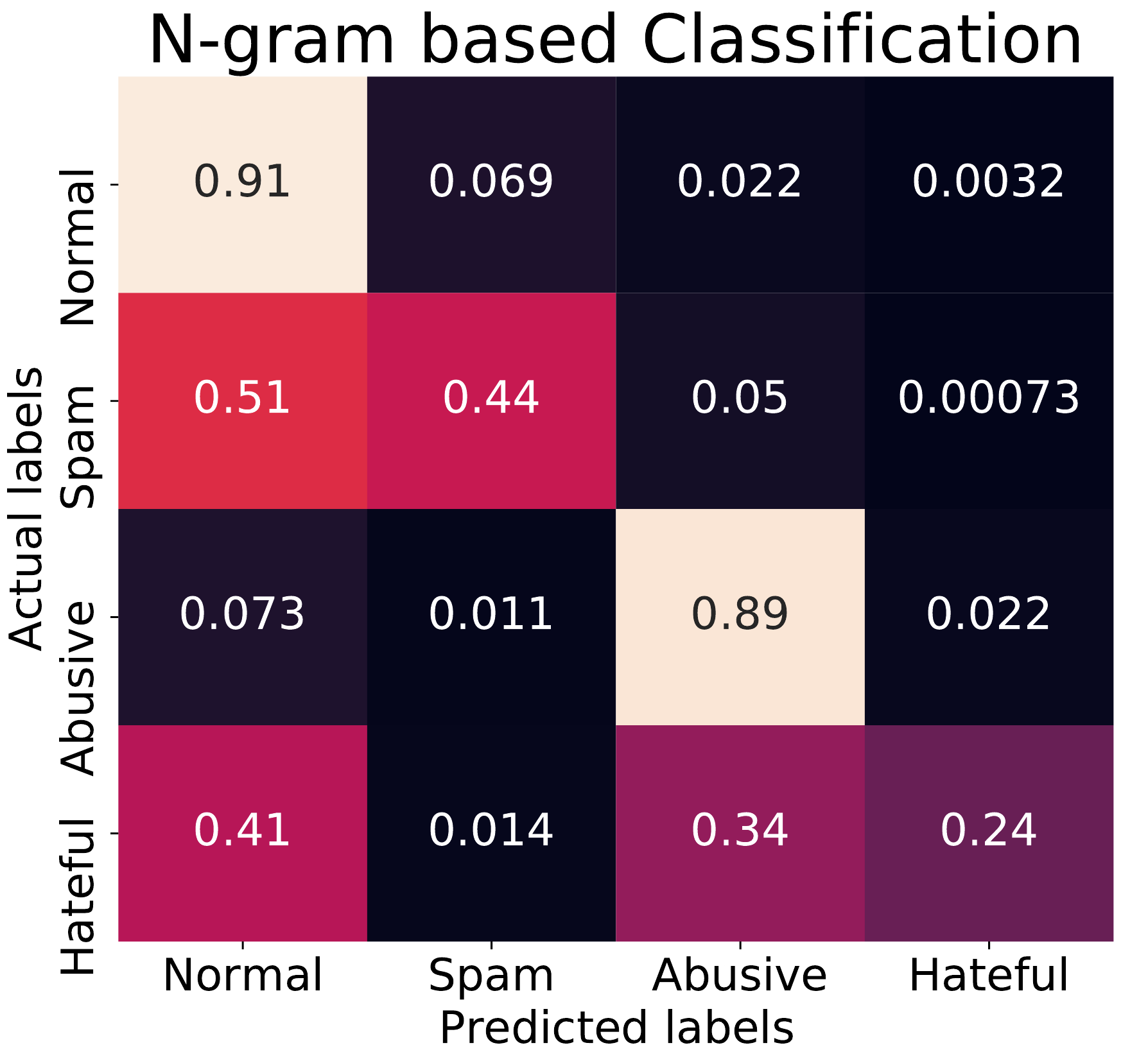}
    \end{minipage}
    \begin{minipage}{0.23\textwidth}
        \centering
        \includegraphics[width=1.0\textwidth]{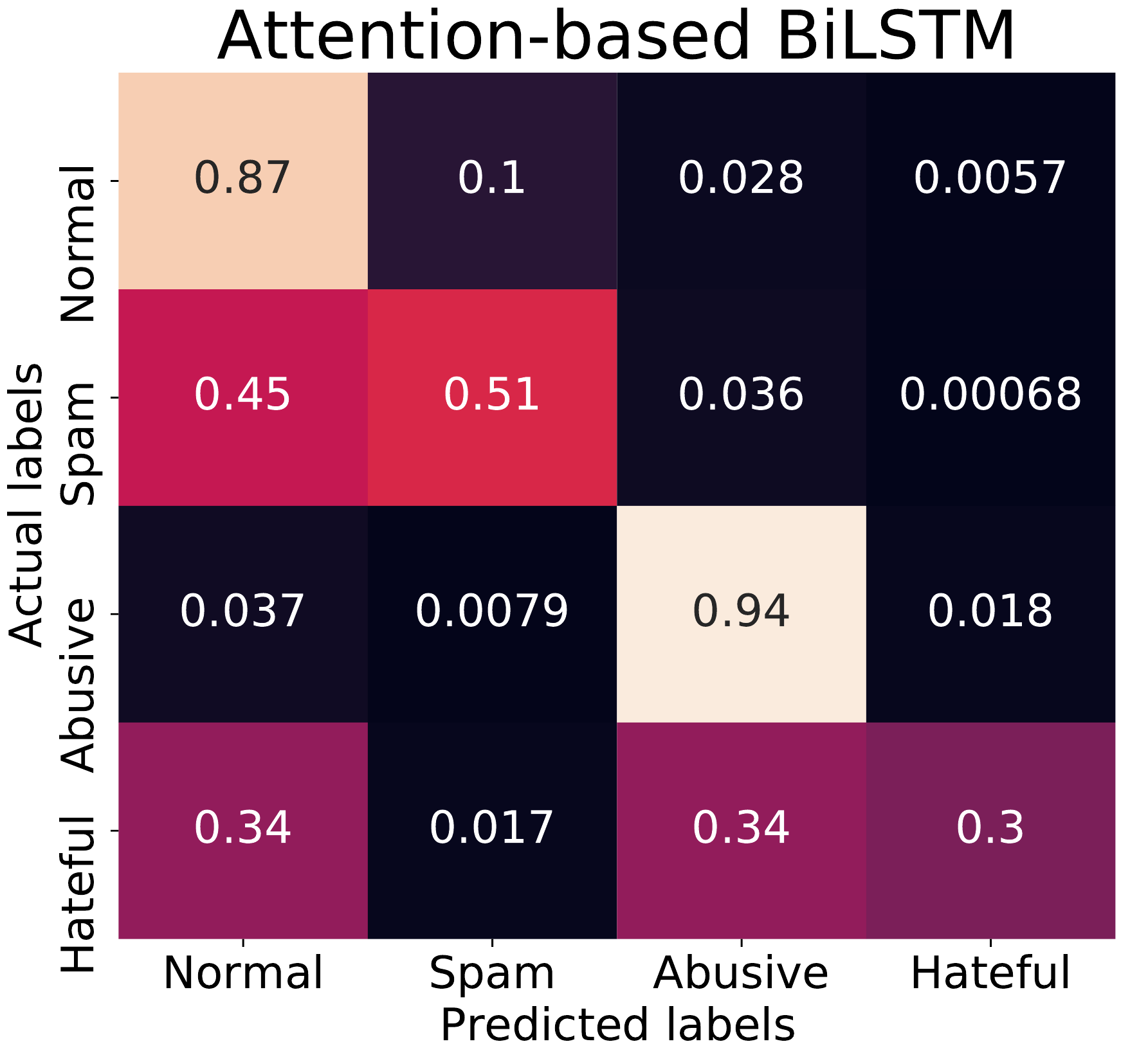}
    \end{minipage}
    \begin{minipage}{0.26\textwidth}
        \centering
        \includegraphics[width=1.0\textwidth]{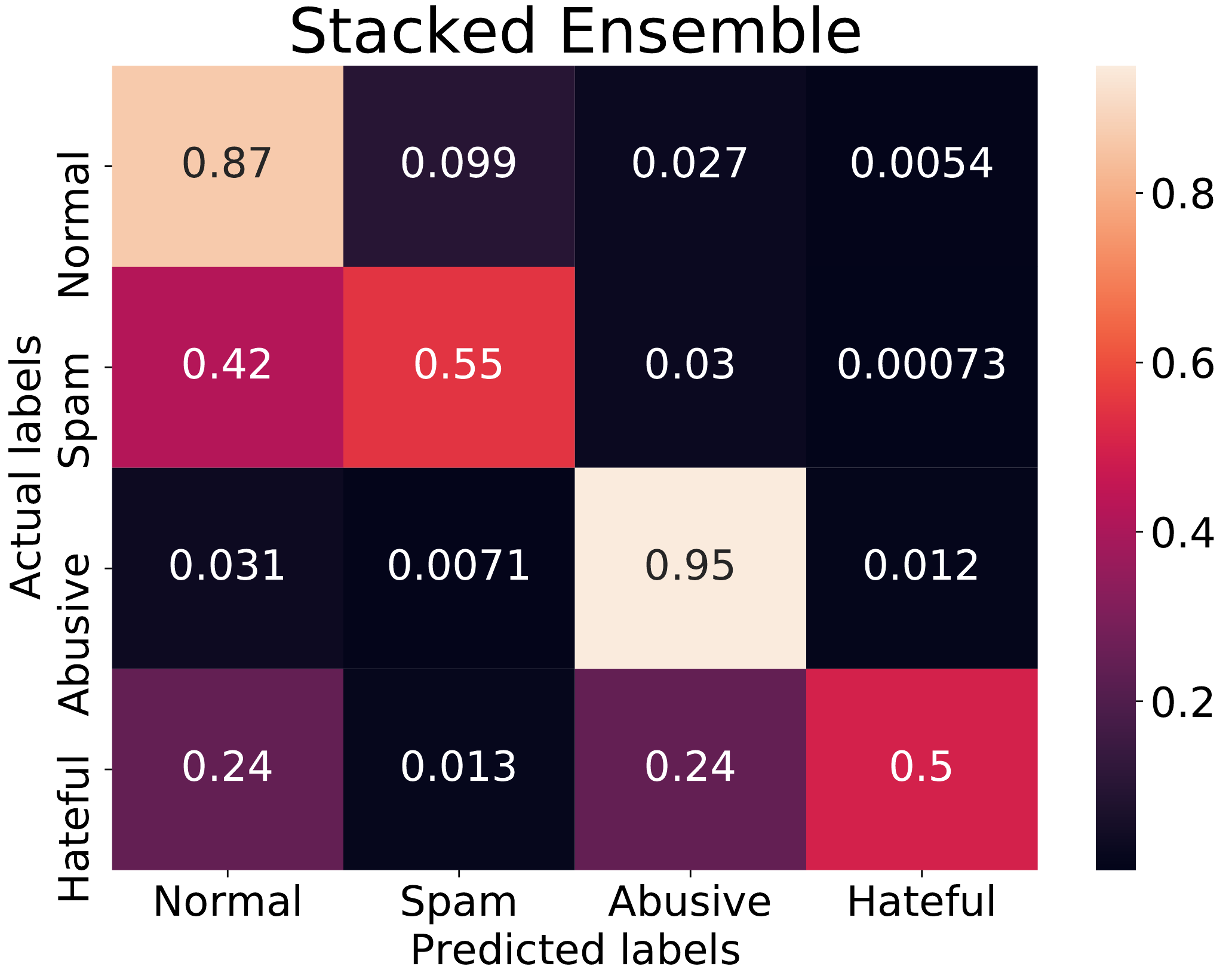}
    \end{minipage}
    \vspace{-2mm}
    \caption{Confusion matrices: the values indicate the fractions of examples in the test set that were classified as any given label.}
    \vspace{-4mm}
    \label{fig:confusion_matrix}
\end{figure*}

Model stacking for ensemble learning involves taking the probability estimations of \textit{base models} and using them as features for training a \textit{meta model}. The general practice is to take diverse and complex base models that make sufficiently different assumptions to solve the predictive task, and then train a simple meta model to ``interpret" these predictions. We treat the above three models as base models and use their predictions over the train (and validation) examples to train (and validate) a logistic regression model (i.e., the meta model). We use the predictions over the test set to evaluate the meta logistic regression model. As we note in Table \ref{tab:classification_results}, the stacked ensemble performs better than all the base models. This reaffirms the diverse modeling assumptions argument that we presented earlier. One consistent shortcoming of the base models, as it is evident from the confusion matrices in Figure \ref{fig:confusion_matrix}, is the tendency to incorrectly classify tweets that are actually \texttt{hateful} as either \texttt{abusive} or \texttt{normal}. It is encouraging to see that the stacked ensemble has a better capability to distinguish between these classes. 

\vspace{-3mm}
\section{Discussion and Future Directions}
This section discusses the presented work in light of the past works and points out some potential directions of future research. We start with a comparison of proposed methods with the existing state-of-the-art, and then discuss some of the shortcomings of our methods. 

\subsubsection{Overall accuracy:}\citeauthor{founta2019unified} (\citeyear{founta2019unified}) proposed a unified deep learning architecture that uses textual, user, and network features to detect abuse. It is encouraging to see (Table \ref{tab:classification_results}) that our stacked ensemble performs on par with their method \textit{without} using \textit{any} user or network features\footnote{\citeauthor{founta2019unified} (\citeyear{founta2019unified}) perform their experiments only on $3$ classes: \texttt{normal}, \texttt{abusive}, and \texttt{hateful} and report $0.84$ accuracy. Whereas, our models, when trained to classify only among these $3$ classes give $\sim 0.89$ accuracy. }. Our methods solely rely on textual properties of the tweets. While this speaks for the competitiveness of our approach, more importantly, it highlights a potential future direction to boost the classification performance of our models. 

\vspace{-1mm}
\subsubsection{Class-wise predictions:} As we mention earlier, the tendency to incorrectly classify \texttt{hateful} tweets as \texttt{abusive} tweets is consistent across all the models. Even though the stacked ensemble mitigates this to a reasonable extent, the problem still persists. Similarly, the tendency to incorrectly classify \texttt{spam} tweets as \texttt{normal} tweets is consistent across all the models; stacked ensemble still being better than others. We believe that the \texttt{spam}-\texttt{normal} confusion can be handled well by incorporating user and network features – spam tweets usually come from bot accounts and few users tend to spam repeatedly.  
However, since the \texttt{hateful}-\texttt{abusive} confusion is central to our discussion around linguistic properties of inappropriate tweets, we present some insights in the following paragraphs. 

\vspace{-2mm}
\subsection{Abusive or Hateful?}
When crowdsourcing the data, \citeauthor{founta2018large} (\citeyear{founta2018large}) found that even though the the label \texttt{hateful} frequently coexists with other labels like  \texttt{abusive}, \texttt{offensive}, and \texttt{aggressive} (all of which are later merged into the \texttt{abusive} category), it is not significantly correlated with any other label. Their definition of hateful tweet emphasizes on the well-defined description of the target groups. As mentioned earlier, this is also indicated in the highly attended words we present in Table \ref{tab:attention_weights}. Owing to this fundamental difference in hateful and abusive tweets, which is largely captured in the linguistic properties of the tweet (as opposed to user or network-related properties), we consider it important to be able to distinguish between these two categories. 

The tendency of our models to confuse between these two categories can be, in part,  attributed to the training data. There is a clear imbalance in data -- hateful tweets only account for $\sim5\%$  of examples. While random oversampling of minority class examples helps in mitigating the consequences of this imbalance, the results are still far from great.  
Furthermore, the average number of annotators that agreed on \texttt{hateful} label for the tweets is lowest when compared to other labels -- i.e., $2.95$ out of $5$ for \texttt{hateful} in comparison to $3.90$, $3.47$, and $3.53$ for \texttt{normal}, \texttt{spam}, and \texttt{abusive}, respectively. This indicates a lack of agreement among the annotators. However, from a linguistic modeling perspective, many hateful tweets are very similar to abusive tweets. For instance, ``i'm fu*king done with twitter'' (\texttt{hateful}) vs ``i'm fu*king done'' (\texttt{abusive}); ``when a n*gga got you fu*ked up web\_link'' (\texttt{hateful}) vs  ``b*tch you got me fu*ked up web\_link'' (\texttt{abusive}); ``some women need to grow the hell up. it's so pathetic.'' (\texttt{hateful}) vs ``some people are so pathetic and need to grow the fu*k up!'' (\texttt{abusive}). These examples further illustrate the point that there's a well-defined target group within hateful tweets and explicitly incorporating that information might be a promising direction for future research.

\vspace{-1mm}
\section{Conclusion}
In this work we discussed three ML models to classify tweets as \texttt{normal}, \texttt{spam}, \texttt{abusive}, and \texttt{hateful}. We then discussed a meta logistic regression model that uses the predictions of these classifiers as features to solve the classification task. Our three base-models provide valuable insights regarding inappropriate tweets: the logistic regression model trained on LIWC features provides insights into psycholinguistic patterns that are emergent in such tweets, the n-gram based classifier provides word embeddings that are tuned with respect to abusive and hateful behavior, and the attention-based BiLSTM highlights some of the important words that influence its predictions. Furthermore, our stacked ensemble provides classification accuracy that is comparable to the state-of-the-art by only using textual properties. Lastly, we discussed the shortcomings of our proposed approaches as well as the subtle linguistic differences in abusive and hateful tweets with a hope that it will influence future research on the topic. 
\vspace{-1mm}
\small{

\bibliographystyle{aaai}
\bibliography{biblio}

\begin{thebibliography}{}

\bibitem[\protect\citeauthoryear{Founta \bgroup et al\mbox.\egroup
  }{2018}]{founta2018large}
Founta, A.~M.; Djouvas, C.; Chatzakou, D.; Leontiadis, I.; Blackburn, J.;
  Stringhini, G.; Vakali, A.; Sirivianos, M.; and Kourtellis, N.
\newblock 2018.
\newblock Large scale crowdsourcing and characterization of twitter abusive
  behavior.
\newblock In {\em ICWSM}.

\bibitem[\protect\citeauthoryear{Founta \bgroup et al\mbox.\egroup
  }{2019}]{founta2019unified}
Founta, A.~M.; Chatzakou, D.; Kourtellis, N.; Blackburn, J.; Vakali, A.; and
  Leontiadis, I.
\newblock 2019.
\newblock A unified deep learning architecture for abuse detection.
\newblock In {\em ACM Conference on Web Science}.

\bibitem[\protect\citeauthoryear{Joulin \bgroup et al\mbox.\egroup
  }{2017}]{joulin2017bag}
Joulin, A.; Grave, {\'E}.; Bojanowski, P.; and Mikolov, T.
\newblock 2017.
\newblock Bag of tricks for efficient text classification.
\newblock In {\em EACL}.

\bibitem[\protect\citeauthoryear{Mikolov \bgroup et al\mbox.\egroup
  }{2013}]{mikolov2013distributed}
Mikolov, T.; Sutskever, I.; Chen, K.; Corrado, G.~S.; and Dean, J.
\newblock 2013.
\newblock Distributed representations of words and phrases and their
  compositionality.
\newblock In {\em NeurIPS}.

\bibitem[\protect\citeauthoryear{Pennebaker, Francis, and
  Booth}{1999}]{liwcPennebaker}
Pennebaker, J.; Francis, M.; and Booth, R.
\newblock 1999.
\newblock Linguistic inquiry and word count (liwc).

\bibitem[\protect\citeauthoryear{Pennington, Socher, and
  Manning}{2014}]{pennington2014glove}
Pennington, J.; Socher, R.; and Manning, C.~D.
\newblock 2014.
\newblock Glove: Global vectors for word representation.
\newblock In {\em EMNLP}.

\bibitem[\protect\citeauthoryear{Sap \bgroup et al\mbox.\egroup
  }{2019}]{sap2019risk}
Sap, M.; Card, D.; Gabriel, S.; Choi, Y.; and Smith, N.~A.
\newblock 2019.
\newblock The risk of racial bias in hate speech detection.
\newblock In {\em ACL}.

\bibitem[\protect\citeauthoryear{Tausczik and
  Pennebaker}{2010}]{tausczik2010psychological}
Tausczik, Y.~R., and Pennebaker, J.~W.
\newblock 2010.
\newblock The psychological meaning of words: Liwc and computerized text
  analysis methods.
\newblock {\em Journal of Language and Social Psychology}.

\bibitem[\protect\citeauthoryear{Zhou \bgroup et al\mbox.\egroup
  }{2016}]{zhou-etal-2016-attention-based}
Zhou, P.; Shi, W.; Tian, J.; Qi, Z.; Li, B.; Hao, H.; and Xu, B.
\newblock 2016.
\newblock Attention-based bidirectional long short-term memory networks for
  relation classification.
\newblock In {\em ACL}.

\end{thebibliography}
}

\end{document}